\documentclass[lettersize,journal]{IEEEtran}
\usepackage{amsmath,amsfonts}
\usepackage{algorithmic}

\usepackage{array}
\usepackage[caption=false,font=normalsize,labelfont=sf,textfont=sf]{subfig}
\usepackage{textcomp}
\usepackage{stfloats}
\usepackage{url}
\usepackage{verbatim}
\usepackage{graphicx}
\usepackage{cite}
\usepackage[linesnumbered,ruled,noline,vlined]{algorithm2e}

\usepackage{multirow}
\usepackage{mathrsfs}
\usepackage{float}
\usepackage{amsfonts,amssymb}
\usepackage{bbm}
\usepackage{booktabs}
\usepackage{mathdots}
\usepackage{amsmath,boxedminipage}
\usepackage[subfigure]{tocloft}
\usepackage{colortbl}
\usepackage{xcolor}
\usepackage{bm}
\usepackage{contour}
\usepackage{tcolorbox}
\usepackage{soul}
\usepackage[colorlinks=true, allcolors=blue]{hyperref}
\usepackage{cleveref}
\crefname{subsection}{Sec.}{}
\crefname{section}{Sec.}{}
\usepackage{cancel}
\usepackage{amsmath}

\begin{document}

\title{Edit-Your-Motion: Space-Time Diffusion Decoupling Learning for Video Motion Editing}

\author{Yi Zuo, 
Lingling Li,~\IEEEmembership{Senior Member,~IEEE}, 
Licheng Jiao,~\IEEEmembership{Fellow,~IEEE}, 
Fang Liu,~\IEEEmembership{Senior Member,~IEEE}, 
Xu Liu,~\IEEEmembership{Member,~IEEE}, 
Wenping Ma,~\IEEEmembership{Member,~IEEE}, 
Shuyuan Yang,~\IEEEmembership{Member,~IEEE}, 
and Yuwei Guo,~\IEEEmembership{Senior Member,~IEEE}
\thanks{The authors are with the Key Laboratory of Intelligent Perception and Image Understanding of the Ministry of Education of China, International Research Center of Intelligent Perception and Computation, School of Artificial Intelligence, Xidian University, Xian 710071, China (e-mail: yiiizuo@163.com; llli@xidian.edu.cn).(Corresponding author: Lingling Li.)}
}

\markboth{Journal of \LaTeX\ Class Files,~Vol.~14, No.~8, August~2021}%
{Shell \MakeLowercase{\textit{et al.}}: A Sample Article Using IEEEtran.cls for IEEE Journals}


\maketitle

\begin{abstract}
        Existing diffusion-based methods have achieved impressive results in human motion editing. However, these methods often exhibit significant ghosting and body distortion in unseen in-the-wild cases. 
        In this paper, we introduce Edit-Your-Motion, a video motion editing method that tackles these challenges through one-shot fine-tuning on unseen cases.  
		Specifically, firstly, we utilized DDIM inversion to initialize the noise, preserving the appearance of the source video and designed a lightweight motion attention adapter module to enhance motion fidelity.
		DDIM inversion aims to obtain the implicit representations by estimating the prediction noise from the source video, which serves as a starting point for the sampling process, ensuring the appearance consistency between the source and edited videos. The Motion Attention Module (MA) enhances the model's motion editing ability by resolving the conflict between the skeleton features and the appearance features.
        Secondly, to effectively decouple motion and appearance of source video, we design a spatio-temporal two-stage learning strategy (STL). In the first stage, we focus on learning temporal features of human motion and propose recurrent causal attention (RCA) to ensure consistency between video frames. In the second stage, we shift focus on learning the appearance features of the source video.
        With Edit-Your-Motion, users can edit the motion of humans in the source video, creating more engaging and diverse content. Extensive qualitative and quantitative experiments, along with user preference studies, show that Edit-Your-Motion outperforms other methods.
\end{abstract}


\section{Introduction}

Deep diffusion models \cite{zhang2023sdmuse, huang2024wavedm} have achieved remarkable success in various fields, including image generation, video generation, and semantic segmentation, due to their powerful generative capabilities. 
This success has also sparked extensive research in video editing, which focuses on controlling the content of source videos using additional conditions (e.g., prompts, depth maps, poses).

Video editing can be categorized into two main directions: video property editing and video motion editing. 
The video attribute editing modifies attributes such as environment color, screen style, and object category based on additional conditions. 
In contrast, the video motion editing focuses on controlling human motion while maintaining the consistency of the human appearance and background with the source video. 
This technology is especially valuable in multimedia applications \cite{koksal2023controllable, fang2024pg}, including advertising, art creation, and movie production. It allows users to effortlessly edit human motion in videos using the video motion model, eliminating the need for complex software tools.

\begin{figure*}
	\centering
	\includegraphics[width=0.9\textwidth]{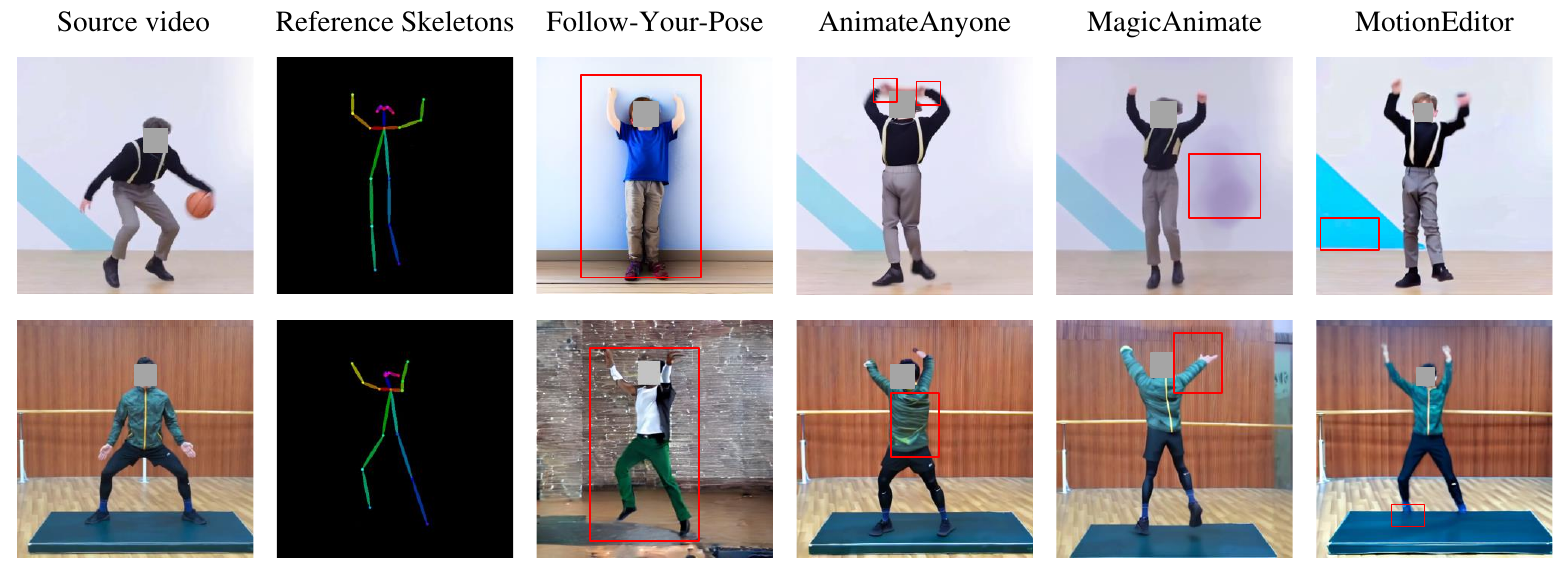}
	\caption{Given the source video (or image) and the reference skeletons, the results generated by the different methods. Red boxes highlight inconsistencies in appearance with the source video, including distortions and ghosting.}
	\label{fig:question1}
\end{figure*}

In previous studies, pose-guided text-to-video methods focus on generating videos with consistent poses using skeleton videos. Motion customization, on the other hand, creates videos with the same motion by learning the motion features from the source video. However, none of them can control the appearance of the human and background in the video.
Xu et al. \cite{xu2024magicanimate} and li et al. \cite{hu2024animate} proposed MagicAnimate and AnimateAnyone, respectively, to control the human motion using a source image/video along with a reference skeleton video.
However, these methods rely on complex human appearance encoders and require training on a large number of video data, making them costly. Additionally, they often exhibit significant ghosting and human distortion in unseen in-the-wild cases. 
This issue arises from the lack of separation between spatial and temporal features, which hinders their ability to quickly adapt to new domains.

MotionEditor \cite{tu2023motioneditor} attempts to address this issue with a two-branch structure. However, since it does not decouple foreground and background during training stage, the foreground and background features overlap in the network. 
As a result, during inference, even with the use of segmentation masks \cite{shan2023incremental,wang2023cal,liu2023exploring}, MotionEditor struggles to effectively decouple the foreground from the background. Fig. \ref{fig:question1} shows the results of Follow-Your-Pose, AnimateAnyone, MagicAnimate and MotionEditor on unseen cases.

In this paper, to overcome the above challenges, we propose Edit-Your-Motion, a video editing method designed to quickly adapt to unseen cases by one-shot fine-tuning. Firstly, we employ DDIM inversion \cite{song2020denoising} to preserve appearance features and design the motion attention module (MA) to enhance the motion editing capability.
The goal of DDIM inversion is to reverse the sampling process, recovering latent noise that retains the appearance information. This latent noise is then used at the start of the sampling process to regenerate the source image.
The motion attention module, composed of self attention, cross attention and temporal attention, enhances the motion editing capabilities of Edit-Your-Motion by optimizing the adaptability between ControlNet \cite{zhang2023adding} and U-Net \cite{ronneberger2015u}.

Secondly, we introduce a spatio-temporal two-stage learning strategy (STL) to enhance the feature extraction capabilities of modules. 
Specifically, in the first stage, we unfreeze the temporal attention layer and motion attention module, while masking the background to concentrate on learning the temporal features of human motion. 
In the second stage, we unfreeze the motion attention module and spatial attention layer to learn the appearance and background features of the source video. Furthermore, we propose recurrent causal attention (RCA) as an alternative to spatial attention. The RCA directly models the relationships ship between frames, ensuring the video consistency.

Overall, Edit-Your-Motion advances the field of video motion editing by quickly adapting to unseen in-the-wild cases while requiring fewer computational resources. Our contributions are summarized as below:

\begin{itemize}
	\item In this paper, we emphasize the importance of decoupled learning in spatio-temporal diffusion models and propose Edit-Your-Motion, a method that enables video motion editing of humans in unseen in-the-wild cases.
	\item We apply DDIM inversion to preserve the appearance features of the source video and introduce a lightweight motion attention module, replacing the traditional two-branch structure to reduce computational costs. At the same time, them improves the consistency of the edited video and the alignment of the motion.
	\item We propose a spatio-temporal decoupling two-stage learning strategy to decouple the spatial and temporal features of the source video, thereby enhancing the feature extraction capabilities of modules.
	\item We conducted experiments on TikTok and unseen in-the-wild videos to demonstrate the robustness and effectiveness of Edit-Your-Motion.
\end{itemize}

\section{RELATED WORK}

\subsection{Pose-guided and Motion Customization Video Generation}

Pose-guided video generation is a method that controls video generation by adding additional human skeletons. 
ControlNet \cite{zhang2023adding} allows for controlled images generation by guiding the Stabel Diffusion \cite{rombach2022high} with various conditional inputs, including Canny edges, Hough lines, segmentation maps, depth maps, etc.
Follow-Your-Pose \cite{ma2024follow} extends this method to video. It controls the videos generation based on human skeletons. It employs a two-stage training strategy to learn poses and maintain temporal consistency.
ControlVideo \cite{zhang2023controlvideo} inherits the structure and weights of ControlNet and adapts it for video by extending self-attention with fully cross-frame attention. Additionally, it introduces a hierarchical sampler to split long videos into separated short video clips with multiple keyframes, 
ensuring long-range coherence by generating keyframes through fully cross-frame attention.
Control-A-Video \cite{zhao2023controlvideo} converts spatial self-attention into key-frame attention and fine-tunes these. Notably, the temporal attention module is used as an additional branch in the diffusion model, passing it through a zero-convolution layer to preserve the output before fine-tuning.

Unlike the pose-guided video generation models, the motion customization video generation models generate videos with the same motion by learning motion features from the source video.
Customize-A-Video \cite{ren2024customize} introduces an Appearance Absorber module to decompose the spatial information of motion, enabling the Temporal LoRA \cite{hu2021lora} to learn the motion details.
MotionCrafter \cite{zhao2023motiondirector} customizes the content and motion by injecting motion information into U-Net's temporal attention module using a parallel spatial-temporal architecture. 
VMC \cite{jeong2023vmc} fine-tunes the temporal attention layer in the video diffusion model to obtain motion vectors for tracking motion trajectories in the target video using residual vectors between consecutive frames, thereby successfully achieving motion customization.

Unlike these methods, video motion editing requires controlling the motion of the source video human while maintaining its appearance and background.

\subsection{Human Image Animate}

With the rapid development of video diffusion modeling, human image animation has attracted the attention of researchers. It aims to transfer motion from a video to a human in a source image, impacting various media industries, including film and advertising.
Some diffusion-based image animation frameworks \cite{karras2023dreampose, wang2023disco} use Stabel Diffusion and ControlNet to implement image animation on reference skeletons. However, they process each video frame independently and ignore the temporal information in the video, resulting in flickering of the generated video.
To address this issue, MagicAnimate \cite{xu2024magicanimate} adds temporal attention to the diffusion model and preserves the appearance and the background using an appearance encoder.
AnimateAnyone \cite{hu2024animate} introduces ReferenceNet, a symmetrical structure to U-Net, to maintain the consistent appearance and capture the spatial details of the source image. Additionally, it includes a pose guider to integrate pose control signals into the denoising process, ensuring precise pose controllability.
To more accurately represent human body changes, Champ \cite{zhu2024champ} uses the 3D parametric human model SMPL \cite{loper2023smpl} as an additional condition to guide U-Net in generating more refined animations.

However, these methods not only require networks that consume substantial GPU memory but also need to be trained on large amounts of video data. Additionally, they often exhibit significant ghosting and distortion in unseen in-the-wild cases.

\subsection{Video Editing}
Video editing allows users make customized changes to source videos, playing a crucial role in multimedia applications such as advertising, art creation, and film production. The current video editing models can be divided into two categories: video property editing \cite{lee2023shape,chai2023stablevideo,wu2023tune,liu2023video,zuo2023cut,bai2024uniedit,qi2023fatezero} and video motion editing \cite{tu2023motioneditor}. 

The goal of the video property editing is to modify the background and appearance (e.g., background, color, shape, etc.) in the source video. Most video attribute editing methods are based on Stabel Diffusion \cite{rombach2022high}, using text prompts to alter the attributes of the source video's content. Text2Live \cite{bar2022text2live} first uses text prompts for video property editing, but because it relies on Layered Neural Atlases \cite{kasten2021layered}, the process is not only time-consuming but also does not guarantee the expected edits. Similarly, \cite{chai2023stablevideo, lee2023shape} use LNA for video editing. 

Tune-A-Video \cite{wu2023tune} presents for the first time One-Shot Video Tuning task, which tunes of the source video in the inflated U-Net. This approach avoids training on large datasets and significantly reduces computational costs. FateZero \cite{qi2023fatezero} discovers that the attentional maps during the inversion effectively preserve structural and motion information. As a result, it utilizes cross-attention to directly fuse these maps into the editing process, and reform self-attention blocks to spatio-temporal attention blocks to enhance the appearance consistency. Then, \cite{li2024video} proposed the Spatio-Temporal Expectation Maximization inversion. Specifically, it utilizes the EM \cite{series1journal} algorithm to iteratively perform E-steps and M-steps to find more compact basis set in the pixels of a video. The low-rank bases are then treated as parameters to be learned.

To maintain time consistency, \cite{peng2023smooth} introduces a simple yet effective noise constraint. The constraint aims to regulate the noise prediction across its temporal neighbors, thus producing a smooth latents and videos. \cite{kara2024rave} employs a noise shuffling strategy that effectively utilizes spatio-temporal interactions between frames. It guides the model to perform spatio-temporal attention to maintain temporal consistency with less computational resources. On the other hand, VidToMe \cite{li2024vidtome} enhances temporal consistency in the generated video by merging attentive tokens across frames.

Unlike video attribute editing, the aim of video motion editing is to align the motion of the source video and the reference skeletons (or reference videos) while maintaining the same background and appearance. MotionEditor \cite{tu2023motioneditor} employs the object's segmentation mask to decouple the appearance and background in the feature layer. The appearance features are then injected into the editing branch to maintain appearance consistency. However, because the object and the background overlap in the feature layer, it is challenging to accurately separate the human appearance from the background features with the segmentation mask.

To address it, we employ DDIM inversion to extract appearance features and design a motion attention module. Additionally, we propose a spatio-temporal two-stage learning strategy to improve feature extraction by decoupling temporal and spatial features over a few training iterations. Moreover, we introduce recurrent causal attention to replace spatial attention, further strengthening inter-frame connections.

\section{METHOD}

\begin{figure*}
	\centering
	\includegraphics[width=1\linewidth]{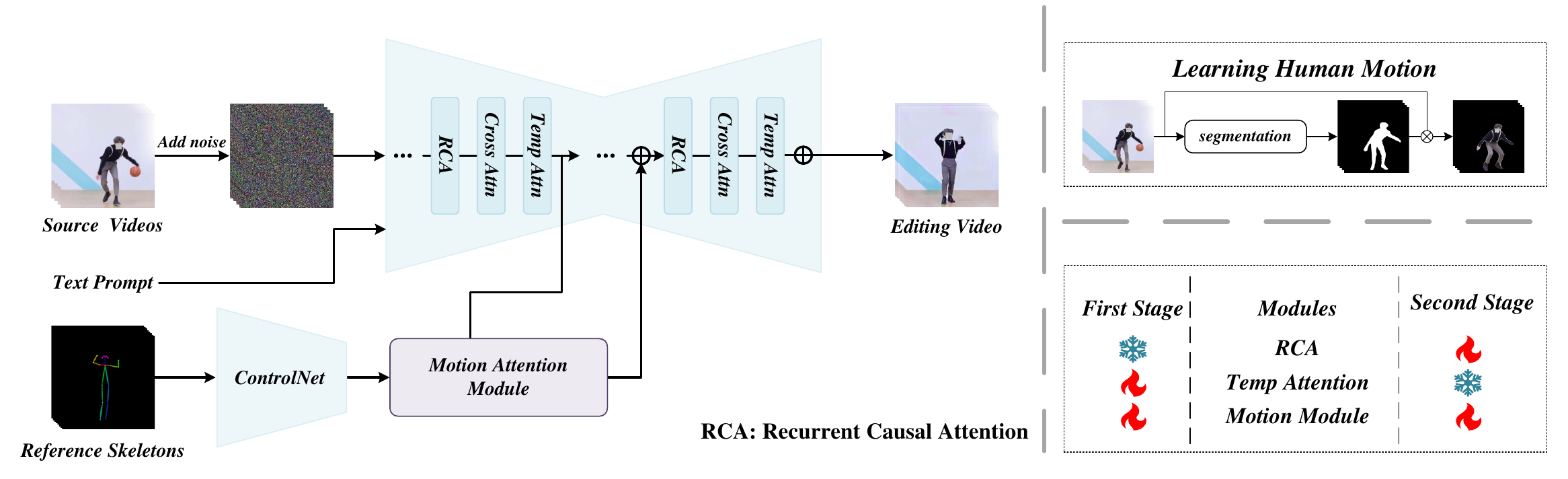}
	\caption{The overall pipeline of Edit-Your-Motion. We employ DDIM inversion to preserve the appearance of the source video and introduce motion attention module to resolve conflicts between skeleton and appearance features. Additionally, we replace spatial attention with recurrent causal attention to enhance inter-frame connections. Finally, to improve the feature extraction capabilities of each module, we design a spatio-temporal decoupling two-stage training strategy that requires only a fewer training iterations.}
	\label{fig:overal_framework}
\end{figure*}

In this paper, we propose Edit-Your-Motion, a method that enables one-shot fine-tuning for unseen in-the-wild cases, ensuring video consistency and fidelity.
To align appearance and motion, we first use DDIM inversion to initialize the noise and design a motion attention module to mitigate the conflict between skeleton and appearance features. 
Second, we design a spatio-temporal decoupling two-stage training strategy to enhance the feature extraction capability of each module with only a few training iterations. 
Furthermore, we propose recurrent causal attention (RCA) as a replacement for spatial attention, allowing spatial features to propagate throughout the video sequence.
Fig. \ref{fig:overal_framework} illustrates the pipeline of Edit-Your-Motion.

\subsection{Preliminaries}\label{subsec:31}

\textbf{Denoising Diffusion Probabilistic Models.}
The denoising diffusion probabilistic models \cite{ho2020denoising, ghosal2023text,yang2023synthesizing,mao2023guided} (DDPMs) consists of a forward diffusion process and a reverse denoising process. During the forward diffusion process, it gradually adds noise $\epsilon$ to a clean image $\boldsymbol{x_0}\sim q(\boldsymbol{x_0})$ with time step $t$, obtaining a noisy sample $x_t$. The process of adding noise can be represented as:
\begin{equation}
	q(\boldsymbol{x}_t|\boldsymbol{x}_{t-1})=\mathcal{N}(\boldsymbol{x}_t|\sqrt{1-\beta_t}\boldsymbol{x}_{t-1},\beta_t\mathbf{I}),
\end{equation}
where $\beta_{t}\in(0,1)$ is a variance schedule. The entire forward process of the diffusion model can be represented as a Markov chain from time $t$ to time $T$,
\begin{equation}
	q(\boldsymbol{x}_{1:T})=q(\boldsymbol{x}_0)\prod_{t=1}^Tq\left(\boldsymbol{x}_t|\boldsymbol{x}_{t-1}\right).
\end{equation}

Then, in reverse processing, noise is removed through a denoising autoencoders $\epsilon_\theta(x_t,t)$ to generate a clean image. The corresponding objective can be simplified to:
\begin{equation}
	L_{DM}=\mathbb{E}_{x,\epsilon\sim\mathcal{N}(0,1),t}\left[\|\epsilon-\epsilon_\theta(x_t,t)\|_2^2\right].
\end{equation}

\noindent\textbf{Latent Diffusion Models.}
Latent Diffusion models (LDM) \cite{rombach2022high,morelli2023ladi,yu2023leveraging} is a newly introduced variant of DDPM that operates in the latent space of the autoencoder. Specifically, the encoder $\mathcal{E}$ compresses the image to latent features $\boldsymbol{z} = \mathcal{E}(\boldsymbol{x})$. Then performs a diffusion process over $z$, and finally reconstructs latent features back into pixel space using the decoder $\mathcal{D}$. The corresponding objective can be represented as:
\begin{equation}
	L_{LDM}=\mathbb{E}_{\mathcal{E}(x),\epsilon\sim\mathcal{N}(0,1),t}\Big[\|\epsilon-\epsilon_\theta(z_t,t)\|_2^2\Big].
\end{equation}

\noindent\textbf{Text-to-Video Diffusion Models.}
Text-to-Video Diffusion Models \cite{song2023relation} train a 3D U-Net $\epsilon^{3D}_{\theta}$ with text prompts $c$ as a condition to generate videos using the T2V model. Given the $F$ frames $\boldsymbol{x}^{1...F}$ of a video, the 3D U-Net is trained by
\begin{equation}
	L_{T2V}=\mathbb{E}_{\mathcal{E}(x^{1...F}),\epsilon\sim\mathcal{N}(0,1),t,c}\left[\left\|\epsilon-\epsilon^{3D}_\theta(z_t^{1...F},t,c)\right\|_2^2\right],
\end{equation}
where $z_t^{1...F}$ is the latent features of $ \boldsymbol{x}^{1...F}$, $z_t^{1...F}=\mathcal{E}(\boldsymbol{x}^{1...F})$.

\subsection{The DDIM Inversion and Motion Attention Module}\label{subsec:32}

\begin{figure}
	\centering
	\includegraphics[width=0.5\textwidth]{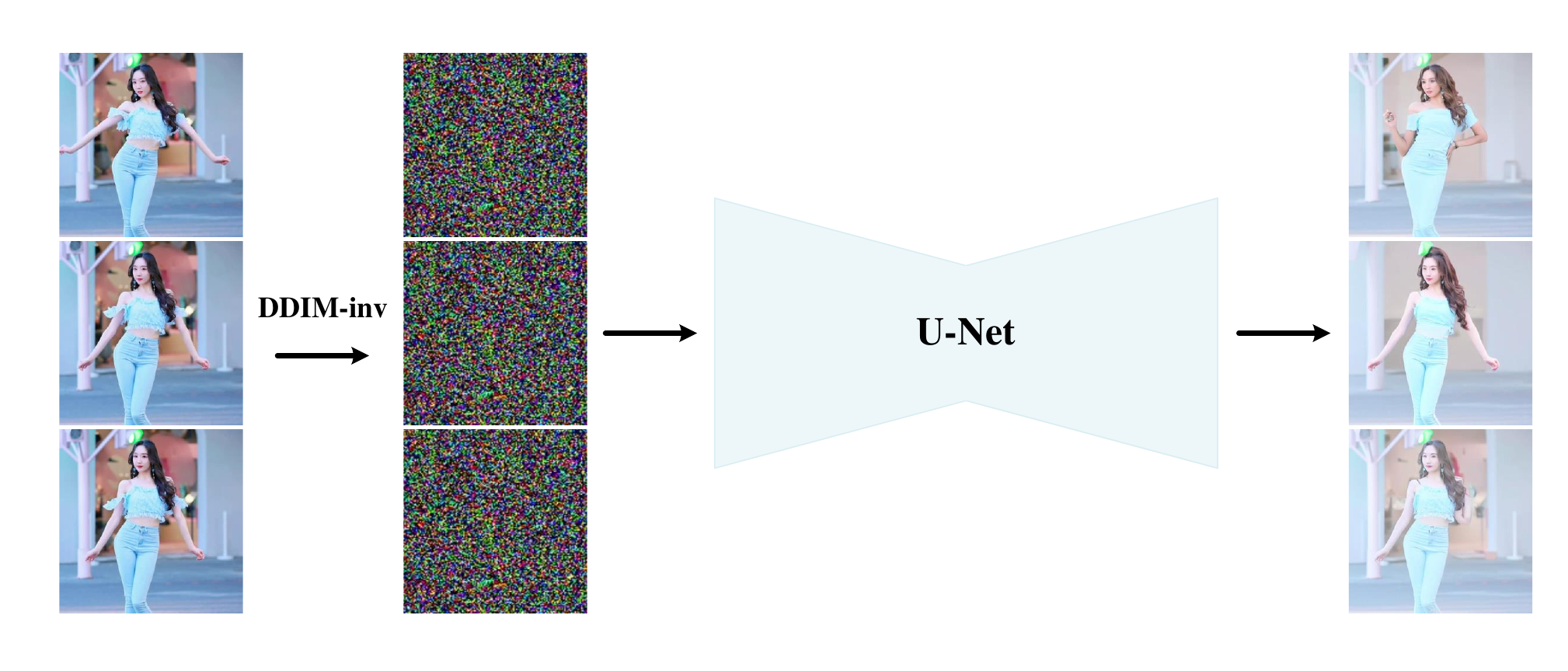}
	\caption{The role of DDIM inversion. The noise obtained by DDIM inversion, directly passed through U-Net, still retains most of the structural features of the source video.}
	\label{fig:structure3}
\end{figure}

The goal of video motion editing is to align the edited video with the appearance of the source video and the motion of the reference skeletons. To achieve this, we use DDIM inversion and propose the motion attention module.

DDIM inversion aims to find a latent noise such that the target image is generated by sampling with this noise as the starting point. Specifically, the DDIM inversion can be described as a gradual addition of noise from $x_0$ to $x_t$, which can be written as:
\begin{equation}
	z_t=\frac{\sqrt{\overline{\alpha_t}}}{\sqrt{\overline{\alpha_t}-1}}(z_{t-1}-\sqrt{1-\alpha_{t-1}^-}\epsilon_t)+\sqrt{1-\overline{\alpha_t}}\epsilon_t
\end{equation}
where $z_t$ represents the lantent feature at time $t$ and $\epsilon_t$ represents the predicted noise. Fig. \ref{fig:structure3} illustrates the process of DDIM inversion.

To enhance the adaptation between ControlNet and U-Net, we propose the motion attention module (MA), which consists of several self attention, cross attention and temporal attention layer.
The self attention layer first models the motion features globally. These features are then integrated with spatial features from the U-Net, allowing skeleton features to match to the spatial features. Finally, temporal attention is applied to ensure inter-frame consistency.

The specific formula for cross attention can be written as follows:
\begin{equation}
	z_t^1 = Attention(Q^c, K^c, V^c),
\end{equation}
\begin{equation}
	z_t^2 = softmax(\dfrac{Linear(z_t^1) (K^u)^T} { \sqrt{d}})V^u 
\end{equation}
where, $c$ represents the ControlNet feature and $u$ represents the U-Net feature.

\subsection{The Recurrent Causal Attention}

\begin{figure}
	\centering
	\includegraphics[width=0.46\textwidth]{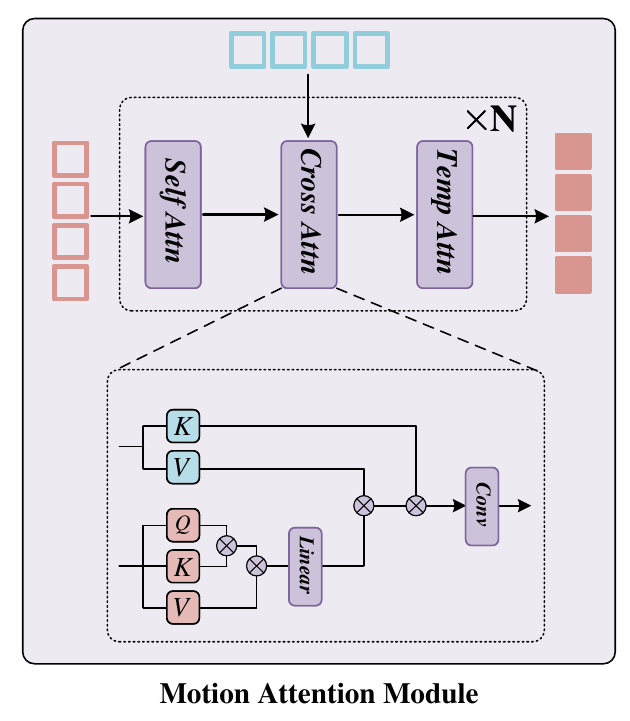}
	\caption{The structure of motion attention module. It consists of a self attention, a cross attention, and a temporal attention, which can mitigate the conflict between skeleton and appearance features extracted by ControlNet and U-Net.}
	\label{fig:structure1}
\end{figure}

We use the inflated U-Net network (spatio-temporal diffusion model) as the backbone of Edit-Your-Motion, consisting of stacked 3D convolutional residual blocks and transform blocks. Each transformer block includes spatial attention, cross attention, temporal attention, and a Feed-Forward network. 

In video, there is a strong relationship between neighboring frames. To help the network better capture this contextual relationship, we propose recurrent causal attention (RCA), an attention module that directly models the contextual dependencies between frames, as illustrated in Fig. \ref{fig:structure1}. Unlike standard causal attention, RCA connects both the previous and next frames, ensuring greater consistency throughout the entire generated video sequence.

In RCA, the key and value are derived by combining the previous frame $z_{v_{i-1}}$ with the next frame $z_{v_{i+1}}$. 
Notably, for the first frame $z_{v_{1}}$, the key and value are obtained from a combination of the first frame $z_{v_{1}}$ and the second frame $z_{v_{2}}$. Similarly, for the final frame $z_{v_{i_{max}}}$, the key and value are derived from the penultimate frame $z_{v_{i_{max}-1}}$ and the final frame $z_{v_{i_{max}}}$.
The formula for RCA is as follows:

\begin{equation}
	Q=W^{Q} z_{v_{i}}, K=W^{K}\left[z_{v_{i-1}}, z_{v_{i+1}}\right], V=W^{V}\left[z_{v_{i-1}}, z_{v_{i}}\right]
\end{equation}
where $\left[\cdot\right]$ denotes concatenation operation. $W^{K}$, $W^{Q}$, $W^{V}$ are the projection matrices.

In summary, compared to standard causal attention, RCA enhances video consistency by establishing a direct connection between the previous and next frames.

\subsection{The Spatio-temporal Two-stage Learning Strategy (STL)}\label{subsec:33}

The goal of diffusion-based video motion editing is to control the motion of the human in the source video based on a reference skeletons video, while maintaining the appearance and background unchanged. 
MagicAnimate and AnimateAnyone achieve human animation by training on large datasets, but they produce ghosting and distortion on unseen in-the-wild cases. 
MotionEditor attempts to address unseen cases through one-shot training, but the incomplete decoupling of motion and appearance leads to low appearance consistency and an inability to accurately align motions.

To tackle this challenge, we introduce the spatio-temporal two-stage learning strategy (STL).
STL is structured into two stages: (1) the first stage focuses on learning the temporal features of human motion, and (2) the second stage emphasizes learning spatial features from appearance and background. Next, we will describe the two stages in detail.

\begin{figure}
	\centering
	\includegraphics[width=0.46\textwidth]{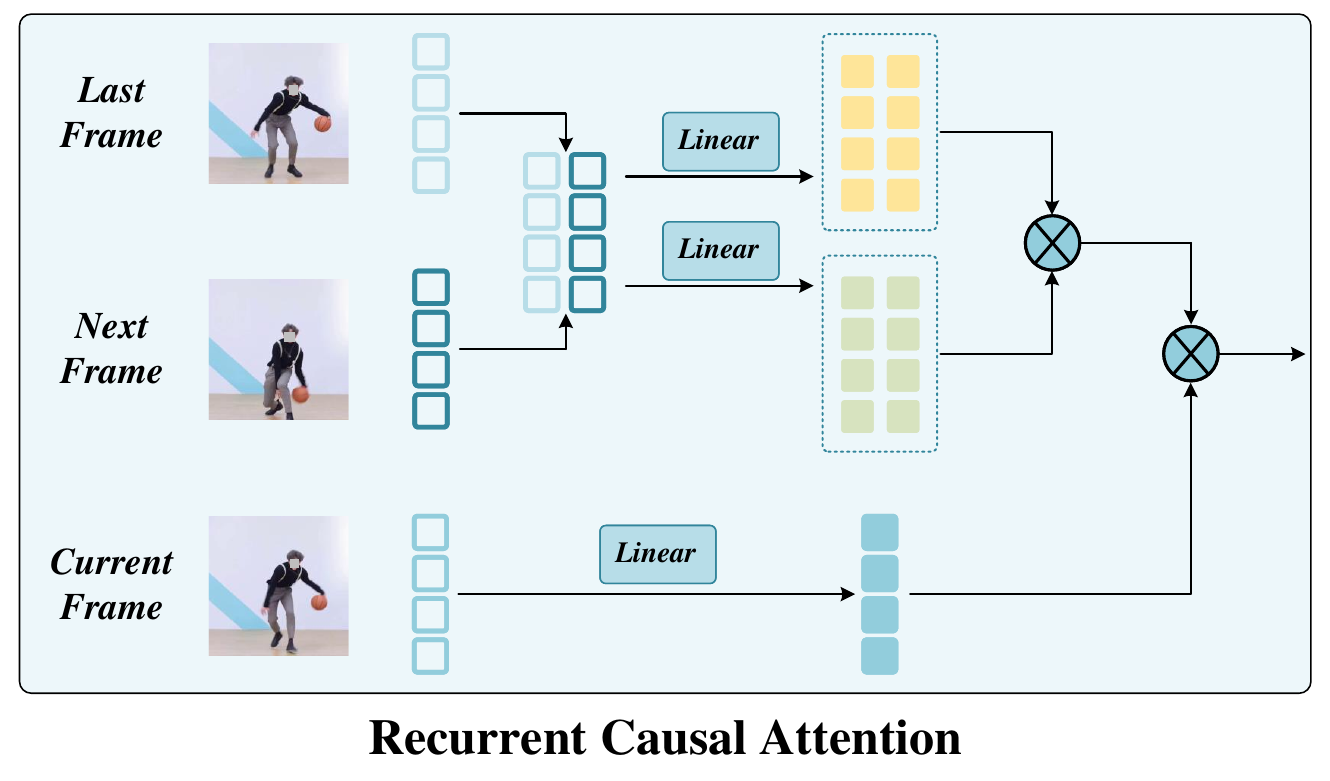}
	\caption{The structure of recurrent causal attention. It directly connects the previous frame to the next, thereby enhancing the consistency of the video.}
	\label{fig:structure2}
\end{figure}

\noindent\textbf{The First Stage: Learning Temporal Features of Human Motion.} In this stage, the spatio-temporal diffusion model focuses on learning the temporal features of human motion. Directly using video frames to train the model would cause an overlap between the features of the human and the background, making it difficult to decouple them later. Therefore, we mask the background and retain only the human appearance. Firstly, we use an existing segmentation network and skeleton extraction network to extract the segmentation masks the skeletons of the human in the video, 
\begin{equation}
	m = \text{network}^\text{s}(v),
\end{equation}
\begin{equation}
	v^m = v \cdot m,
\end{equation}
\begin{equation}
	v_{sk}=\text{network}^\text{k}(v^m),
\end{equation}
where $\text{network}^\text{s}$ represents the segmentation network, $\text{network}^\text{k}$ is the skeleton extraction network, $v_{sk}$ refers to the human skeletons, and $v^m$ is the video that retains only the appearance of the human. 

Secondly, we obtain the latent features and add noise to initialize the inputs to the diffusion model,
\begin{equation}
	z_{t}^{m} = \text{addnoise}(\mathcal{E}(v^m), \epsilon, timestep),
\end{equation}
where $z_{t}^{m}$ is the latent features, $\epsilon$ is random noise, $timestep$ is time step and $\mathcal{E}(\cdot)$ is the VAE encoder.
Then, we input $v_{sk}$ into ControlNet and motion attention module,
\begin{equation}
	f_{cn}=\text{ControlNet}(v_{sk}, p),
\end{equation}
\begin{equation}
	f_{sk}=\text{MA}(f_{cn}, z_t^u),
\end{equation}
where $f_{cn}$ represents the skeleton features from ControlNet, $f_{sk}$ represents the features processed by the motion attention module, and $z_t^u$ represents the spatial features from U-Net. 

Finally, we train the diffusion model by freezing the other parameters and activating only the temporal attention and motion attention modules. The loss function is defined as:
\begin{equation}
	L=\mathbb{E}_{z_{t}^{m},\epsilon\sim\mathcal{N}(0,1),t,p,f_{sk},f_{v}}\left[\left\|\epsilon-\epsilon^{unet}_\theta(z_{t}^{m},t,p,f_{sk})\right\|_2^2\right].
\end{equation}

\noindent\textbf{The Second Stage: Learning Spatial Features from Appearance and Background.} Unlike the first training stage, in the second stage, we do not occlude the background, allowing the spatio-temporal diffusion model to learn spatial features from both the appearance and background of the video. Specifically, we activate the recurrent causal attention and motion attention module, while freezing the other parameters.

\begin{algorithm}[tbp]
	\caption{Skeleton Offset Algorithm}\label{algorithm1}
	\KwIn{source skeletons $v^s_{sk}$, reference skeletons $v^r_{sk}$.}
	\KwOut{offset reference skeletons $v^{ro}_{sk}$.}
	Get skeletons center position: $p^s = GetCenter(v^s_{sk})$\;
	Get skeletons center position: $p^r = GetCenter(v^r_{sk})$\;
	Get offset distance: $dis = p^r-p^s$\;
	Get the height and width of the skeleton: $h^{s}, w^{s} = GetHW(v^s_{sk})$\;
	Get the height and width of the skeleton: $h^{r}, w^{r} = GetHW(v^r_{sk})$\;
	$scale_h=h^{s}/h^{r}$\;
	$scale_w=w^{s}/w^{r}$\;
	\For{$p^r_{sk} \in v^r_{sk}$}{
		$(x^r_{sk}, y^r_{sk})\in p^r_{sk}$, $(x^r, y^r)\in p^r$, $(x^d, y^d)\in dis$\;
	  	$x^{ro}_{sk}=x^r+(x^r_{sk}-x^r)*scale_w-x^d$\;
		$y^{ro}_{sk}=y^r+(y^r_{sk}-y^r)*scale_h-y^d$\;
		$p^{ro}_{sk}\in (x^{ro}_{sk}, y^{ro}_{sk})$\;
		$v^{ro}_{sk} \leftarrow p^{ro}_{sk}$\;
	}
\end{algorithm}

\noindent\textbf{The Inference Stage.} To preserve the spatial and temporal information of the source video as much as possible, we perform DDIM inversion on the source images/video to obtain latent noise $z^*$, 
\begin{equation}
	z^* = \text{DDIM-inv}(\mathcal{E}(\boldsymbol{x})).
\end{equation}
where $\text{DDIM-inv}$ represent DDIM inversion. As shown in Fig. \ref{fig:structure3}, this approach ensures maximum preservation of the structure of the source video.

It is worth noting that there may be an offset between the position of the reference skeletons and the position of the human in the source video, which can increase the likelihood of ghosting in the generated edited video. To mitigate this, we adjust the position of the reference skeleton by the skeleton offset algorithm, as described in Algorithm \ref{algorithm1}.

\section{EXPERIMENTAL}
\begin{figure*}
	\centering
	\includegraphics[width=1\linewidth]{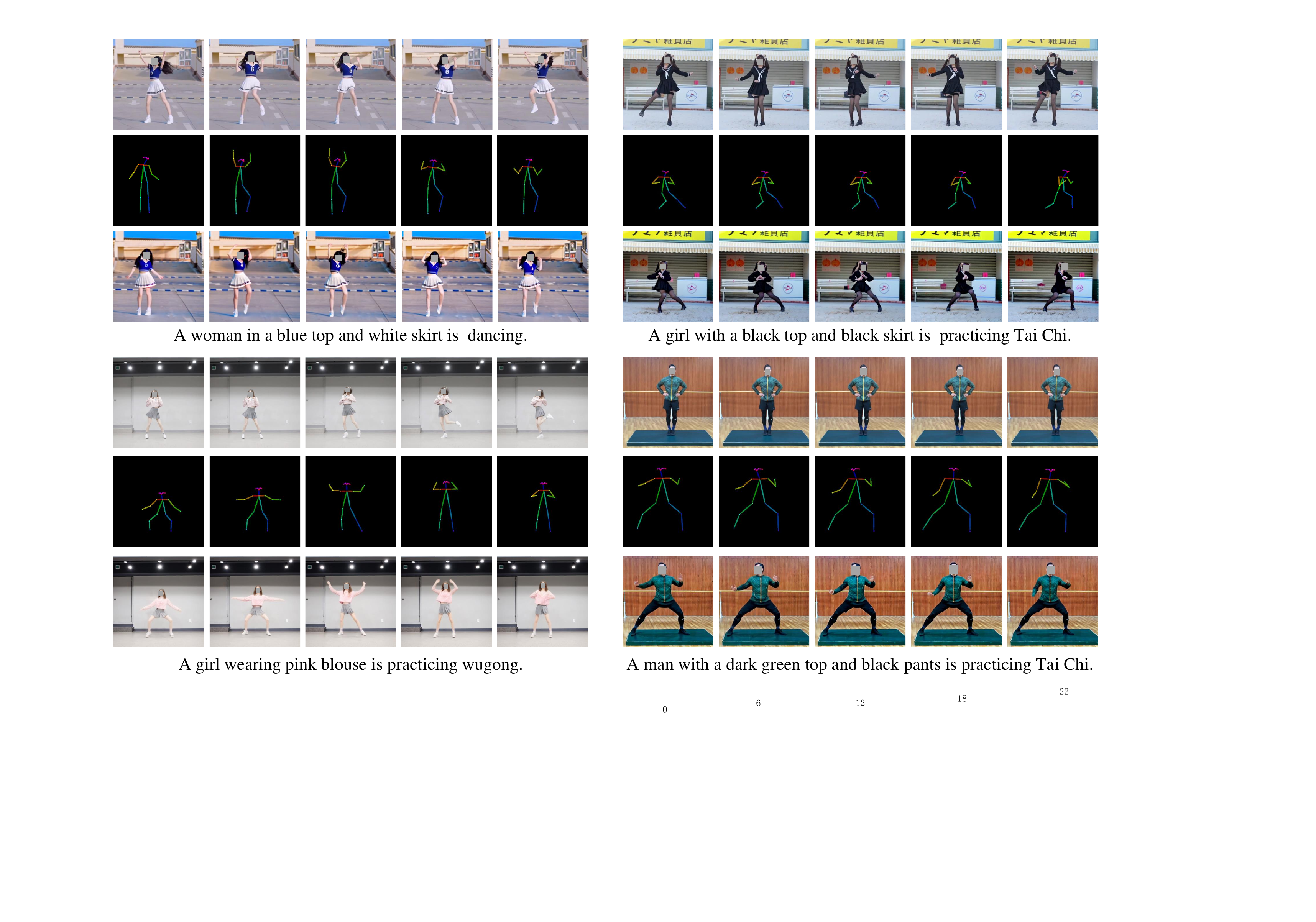}
	\caption{Some examples of motion editing results for Edit-Your-Motion in the unseen in-the-wild cases. Edit-Your-Motion not only adapts to both outdoor and indoor scenes but also facilitates motion editing for a variety of movements, such as dancing, wugong, and Tai Chi.}
	\label{fig:visual_results_sample}
\end{figure*}

\subsection{Datasets and Experimental Setup}

We conduct our evaluation on the Tiktok \cite{jafarian2021learning} and YouTube datasets, following the previous works \cite{wu2023tune, wang2023disco, hu2024animate, xu2024magicanimate, zhu2024champ, tu2023motioneditor}. 

\begin{itemize}
	\item \textbf{The TikTok dataset:} It contains 340 video sequences, along with the corresponding masks and sensepose annotations.
	\item \textbf{The unseen in-the-wild cases:} we collected 80 video sequences, each with at least 70 frames, from YouTube to evaluate the robustness of the model as unseen data.
\end{itemize}

Our proposed Edit-Your-Motion is based on the Latent Diffusion Model \cite{rombach2022high} (Stabel Diffusion). For each unseen in-the-wild case, we fine-tune the model for 300 iterations in each of the two training stages, using a learning rate of $3 \times 10^{-5}$. For inference, we used the DDIM sampler \cite{song2020denoising} with no classifier guidance \cite{ho2022classifier}. All methods are implemented on PyTorch using NVIDIA A100 GPU.

\subsection{Comparisons Method}
\begin{table*}[htbp]
	\centering
	\caption{The difference between other comparison methods and video motion editing.}
	  \begin{tabular}{ccr}
	  \toprule
	  Related work & Method & \multicolumn{1}{c}{Constraint} \\
	  \midrule
	  Motion customization & MotionDirector & Cannot control generated backgrounds and human appearance \\
	  Pose-guided video generation & Follow-Your-Pose & Cannot control generated backgrounds and human appearance \\
	  Video attribute editing & Tune-A-Video & Cannot edit human motion \\
	  Image animation & Champ, AnimateAnyone and MagicAnimate & Large training cost, poor performance on unseen cases \\
	  \bottomrule
	  \end{tabular}%
	\label{tab:diff_vme}%
\end{table*}%
  
Since there are fewer methods specifically for video motion editing, to demonstrate the superiority of Edit-Your-Motion, we select the state-of-the-art methods from related fields, including motion customization, pose-guided video generation, video attribute editing, video motion editing, and image animation, for comparison. 
There are a total of seven methods used for comparison, including: MotionDirector \cite{zhao2023motiondirector}, Follow-Your-Pose \cite{ma2024follow}, Tune-A-Video \cite{wu2023tune}, MagicAnimate \cite{xu2024magicanimate}, AnimateAnyone \cite{hu2024animate}, Champ \cite{zhu2024champ} and MotionEditor \cite{tu2023motioneditor}. 
The difference, between these comparison methods and video motion editing, can be seen in Table \ref{tab:diff_vme}. Additionally, since the existing methods for motion customization, pose-guided video generation, and video attribute editing were not specifically designed for video motion editing, we appropriately modified these methods for comparison. Descriptions of all the compared methods are provided below:

\begin{itemize}
	\item \textbf{MotionDirector} \cite{zhao2023motiondirector} generates motion-aligned videos by decoupling appearance and motion in reference videos for video motion customization. To adapt it for the video motion editing task, during the training stage, we train MotionDirector to learn the motion of the reference video.
	\item \textbf{Follow-Your-Pose} \cite{ma2024follow} generates pose controllable videos using two-stage training. We first take the latent of the source video and invert them using DDIM tp initialize the noise. Then, the skeleton from the reference video is input into Follow-Your-Pose to generate the edited video.
	\item \textbf{Tune-A-Video} \cite{wu2023tune} is the first method to introduce one-shot video editing. It inflates a pre-trained T2I diffusion model to 3D to handle the video task. We input the skeleton of the reference video into ControlNet and use the output of ControlNet as an additional condition to guide the alignment of the motion.
	\item \textbf{MotionEditor} \cite{tu2023motioneditor} is the first to explore video motion editing while ensuring that the human appearance and background remain unchanged.
	\item \textbf{AnimateAnyone} \cite{hu2024animate} introduces a symmetrically designed ReferenceNet alongside U-Net to maintain the appearance consistency. Additionally, it integrates temporal attention into the denoising U-Net to maintain temporal consistency.
	\item \textbf{MagicAnimate} \cite{xu2024magicanimate} incorporates temporal attention blocks into diffusion networks to encode temporal information. Additionally, it introduces appearance encoders specifically designed to preserve human and background features.
	\item \textbf{Champ} \cite{zhu2024champ} utilizes the 3D parametric human model, SMPL \cite{loper2023smpl}, to provide a unified representation of the human body variations, allowing for more refined human animation.
\end{itemize}

\begin{figure*}
	\centering
	\includegraphics[width=1\linewidth]{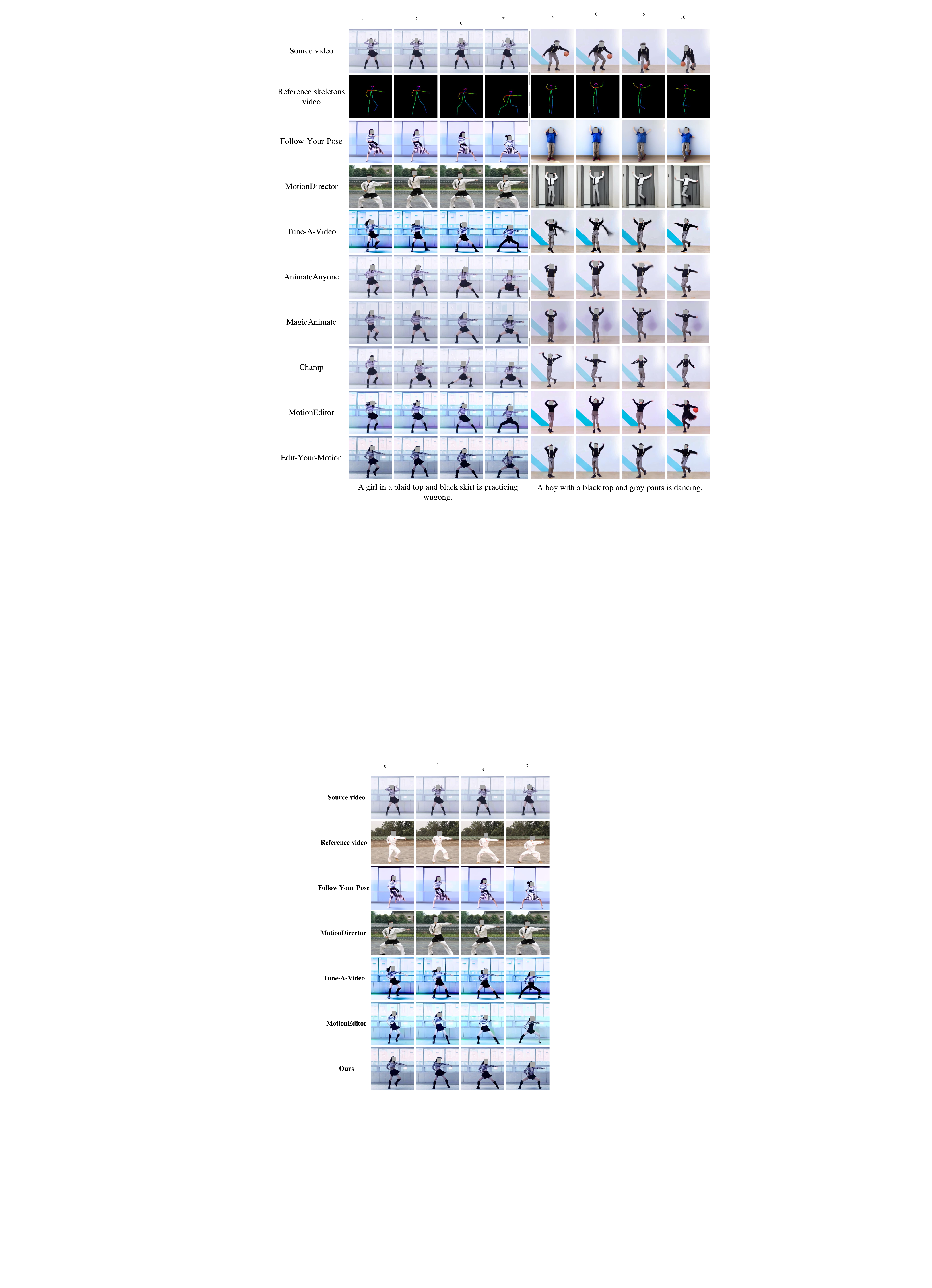}
	\caption{Qualitative comparison results of Follow-Your-Pose, MotionDirector, Tune-A-Video, AnimateAnyone, MagicAnimate, Champ, MotionEditor and Edit-Your-Motion. The results demonstrate that Edit-Your-Motion not only successfully achieves motion alignment with the reference skeletons video but also maintains consistency in the background and human, resulting in high-quality video output compared to other methods.}
	\label{fig:visual_comparison}
\end{figure*}

\begin{table*}[htbp]
	\centering
	\caption{Quantitative comparison of MagicAnimate, AnimateAnyone, Champ and our proposed Edit-Your-Motion on the TikTok benchmark. The highest score is marked in \textbf{BOLD}.}
	\begin{tabular}{ccccccc}
		\toprule
		Method & L1 $\downarrow$ & PSNR $\uparrow$ & SSIM $\uparrow$ & LPIPS $\downarrow$ & FID-VID $\downarrow$ & FVD $\downarrow$ \\
		\midrule
		MagicAnimate \cite{xu2024magicanimate} & 3.13E-04 & 29.16 & 0.714 & 0.239 & 21.75 & 179.07 \\
		AnimateAnyone \cite{hu2024animate} & - & 29.56 & 0.718 & 0.285 & - & 171.9 \\
		Champ \cite{zhu2024champ} & 2.94E-04 & \textbf{29.91} & 0.802 &  0.234 & 21.07 & \textbf{160.82}\\
		\midrule
		Ours & \textbf{2.81E-05} & 29.33 & \textbf{0.813} & \textbf{0.166} & \textbf{20.95} & 166.49\\
		\bottomrule
	\end{tabular}%
	\label{tab:quantitative_comparison_tiktok}%
\end{table*}%

\begin{table*}[htbp]
	\centering
	\caption{Quantitative comparison of MagicAnimate, AnimateAnyone, Champ and our proposed Edit-Your-Motion on in-the-wild cases. The highest score is marked in \textbf{BOLD}.}
	\begin{tabular}{ccccccc}
		\toprule
		Method & L1 $\downarrow$ & PSNR $\uparrow$ & SSIM $\uparrow$ & LPIPS $\downarrow$ & FID-VID $\downarrow$ & FVD $\downarrow$ \\
		\midrule
		MagicAnimate \cite{xu2024magicanimate} & 5.64E-05 & 21.66 & 0.838 & 0.140 & 36.35 & 629.85 \\
		AnimateAnyone \cite{hu2024animate} & 4.86E-05 & 20.90 & 0.796 & 0.159 & 31.38 & 728.87 \\
		Champ \cite{zhu2024champ} & 5.75E-05 & 19.05 & 0.752 &  0.206 & 36.23 & 960.35\\
		\midrule
		Ours & \textbf{4.59E-05} & \textbf{23.03} & \textbf{0.846} & \textbf{0.131} & \textbf{31.19} & \textbf{562.34}\\
		\bottomrule
	\end{tabular}%
	\label{tab:quantitative_comparison_wild}%
\end{table*}%
\begin{figure*}
	\centering
	\includegraphics[width=1\linewidth]{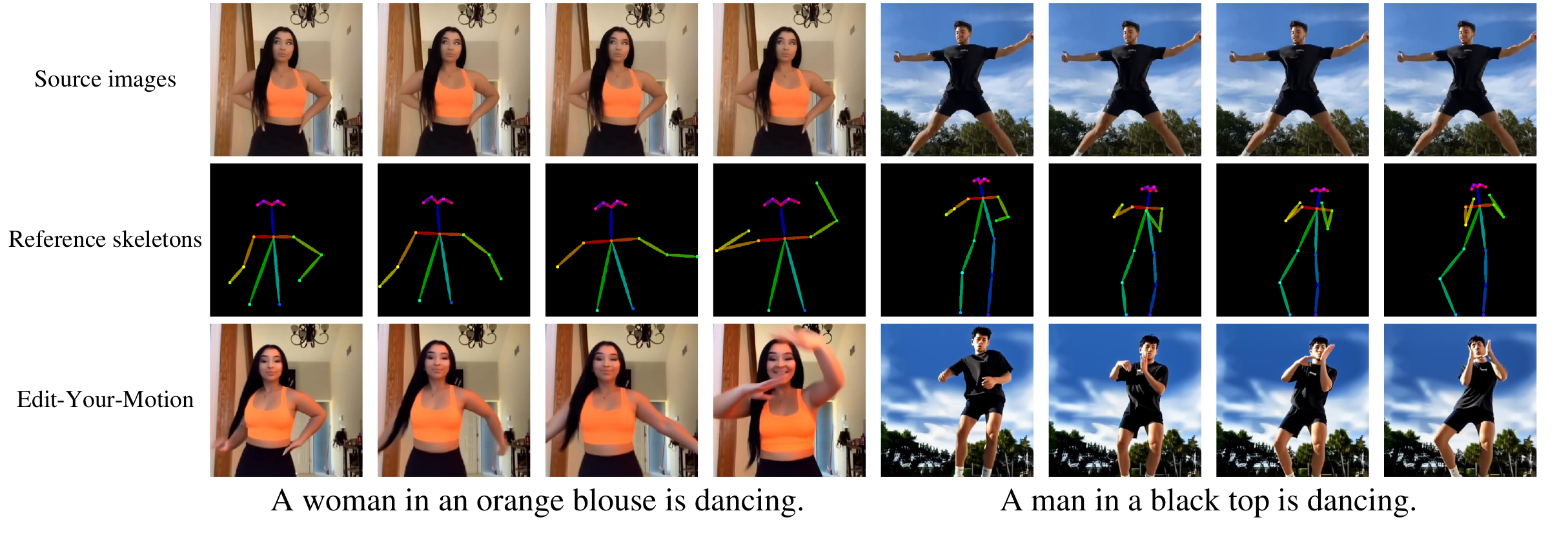}
	\caption{Some examples of motion editing results for Edit-Your-Motion in TikTok dataset. Edit-Your-Motion not only aligns motion and appearance but also ensures a high inter-frame consistency.}
	\label{fig:visual_results_sample2}
\end{figure*}

\begin{table}[htbp]
	\centering
	\caption{Ablation study of RCA, motion attention module and STL in unseen in-the-wild cases. "w/o RCA" indicates that spatial attention is not replaced. "w/o MA" indicates that the motion attention module is not utilized and ControlNet features are directly input into U-Net. "w/o STL" indicates that all modules are trained in one training stage.}
	\scalebox{1}{
	\begin{tabular}{ccccccc}
		\toprule
		Method & PSNR $\uparrow$ & SSIM $\uparrow$ & LPIPS $\downarrow$ & FVD $\downarrow$ \\
		\midrule
		w/o RCA & 22.57 & 0.833 & 0.162 & 602.17  \\
		w/o MA & 21.36 & 0.781 & 0.195 & 562.34\\
		w/o STL & 20.24 & 0.813 & 0.162 & 638.17\\
		Ours & 23.03 & 0.846 & 0.131 & 562.34\\
		\bottomrule
	\end{tabular}%
	}
	\label{tab:ablation_study}%
\end{table}%

\subsection{Evaluation}
\noindent \textbf{Visualization}. Our method effectively aligns the motion of the human in the source video with the reference skeletons, even in unseen in-the-wild cases. 
Fig. \ref{fig:visual_comparison} presents the results of the comparison between Edit-Your-motion and other methods. As shown, Follow-Your-Pose and MotionDirector fail to maintain the appearance and background consistency with the source video. 
Video generated by Tune-A-Video, AnimateAnyone, MagicAnimate, and Champ suffers from distortion and artifacts. MotionEditor generates video with inconsistent human appearance between frames and noticeable differences from the source video. 
Our proposed Edit-Your-Motion accurately controls the motion while preserving both the human appearance and background. Fig. \ref{fig:visual_results_sample} and \ref{fig:visual_results_sample2} provide more examples of Edit-Your-Motion's visualization on the unseen data and TikTok dataset. It is worth noting that if only one source image is available, we repeat it $N$ times to create a video of length $N$, where $N$ matches the length as the reference skeleton video.

\noindent \textbf{Sensitivity Study}
\begin{table}[htbp]
	\centering
	\caption{Sensitivity Study. In the unseen data, the PSNR varies under different combinations of the number of two-stage fine-tuning of the STL.}
	  \begin{tabular}{c|c|cccc}
	  \toprule
	  \multicolumn{2}{c|}{\multirow{2}[4]{*}{Fine-tuning iterations}} & \multicolumn{4}{c}{Stage2} \\
  \cmidrule{3-6}    \multicolumn{2}{c|}{} & 100   & 200   & 300   & 400 \\
	  \midrule
	  \multirow{4}[2]{*}{Stage1} & 100 & 20.55 & 21.43 & 20.98 & 20.72 \\
			& 200 & 20.79 & 21.26 & 22.55 & 22.56 \\
			& 300 & 21.05 & 22.57 & 23.03 & 22.92 \\
			& 400 & 21.17 & 22.04 & 22.79 & 22.87 \\
	  \bottomrule
	  \end{tabular}%
	\label{tab:sensitivity}%
  \end{table}%
We conducted sensitivity experiments on the number of fine-tuning iterations in the two-stage learning strategy (STL) to assess the impact on PSNR for Edit-Your-Motion in unseen in-the-wild cases. The results are presented in Table \ref{tab:sensitivity}. Our findings indicate that the PSNR is highest when the fine-tuning iterations are set to 300 for both stages.

\noindent \textbf{TikTok benchmark} \cite{jafarian2021learning}. The TikTok dataset comprises 340 video sequences and serves as the benchmark for image animation. To evaluate the quality of image generation, we used SSIM \cite{wang2004image}, LPIPS \cite{zhang2018unreasonable}, PSNR \cite{hore2010image} and L1 as metrics. For video evaluation, we employed FID-VID \cite{balaji2019conditional} and FVD \cite{unterthiner2018towards} as metrics.
Table \ref{tab:quantitative_comparison_tiktok} presents the results of our quantitative comparison on the TiTok benchmark. Edit-Your-Motion outperforms the other methods in the L1, SSIM, LPIPS, and FID-VID metrics. This indicates that our method produces videos with higher consistency and superior reconstruction quality.

\noindent \textbf{The Unseen In-the-wild Cases}. For MagicAnimate, AnimateAnyone and Champ, we still used SSIM, LISPIS, PSNR, L1, FID-VID and FVD as metrics, as shown in Table \ref{tab:quantitative_comparison_wild}. The results demonstrate that our method is better adapted to unseen in-the-wild data compared to these methods. 

For Follow-Your-Pose, MotionDirector, Tune-A-Video, and MotionEditor, we employed the Text Alignment, Temporal Consistency, LPIPS-N, and LPIPS-S metrics that were used in their papers. (1) Text Alignment (TA). We use CLIP \cite{radford2021learning} to compute the average cosine similarity between the prompt and the generated frames. (2) Temporal Consistency (TC). We use CLIP to obtain image features and compute the average cosine similarity between neighbouring video frames. (3) LPIPS-N (L-N): We calculate Learned Perceptual Image Patch Similarity \cite{zhang2018unreasonable} between generated neighbouring frames. (4) LPIPS-S (L-S): We calculate Learned Perceptual Image Patch Similarity between generated frames and source frames. 

In addition, we invited 70 participants to take part in the user study. Each participant could see the source video, the reference skeletons video, and the results from both our method and the other comparison methods. For each case, we combined the results of Edit-Your-Motion with those of each of the four comparison methods. Then, we posed three questions to evaluate Text Alignment, Content Alignment and Motion Alignment. The three questions are: "Which is more aligned to the text prompt?", "Which is more content aligned to the source video?" and "Which is more motion aligned to the reference skeletons video?". Table \ref{tab:comparison_wild} presents the results of the user study. This score was calculated using the following formula: 
\begin{equation}
	score=\dfrac{count_{our}} {count_{our}+count_{com}},
\end{equation}
where $count_{our}$ represents the number of users who selected Edit-Your-Motion, and $count_{com}$ denotes the number of users who chose one of the comparison methods (Follow-Your-Pose, MotionDirector, Tune-A-Video, and MotionEditor). 
The higher the $score$, the greater the user agreement with Edit-Your-Motion. For example, 76.43\% of participants believe that Edit-Your-Motion has better text alignment than MotionEditor, 82.15\% feel that Edit-Your-Motion has better content alignment than Tune-A-Video, and 86.18\% agree that Edit-Your-Motion has better motion alignment than MotionDirector.

\begin{table*}[htbp]
  \centering
  \caption{Quantitative comparison and user study of Follow-Your-Pose, MotionDirector, Tune-A-Video, MotionEditor and our proposed Edit-Your-Motion on in-the-wild cases. The highest score is marked in \textbf{BOLD}.}
    \begin{tabular}{c|cccc|rrr}
    \toprule
    \multirow{2}[3]{*}{Method} & \multicolumn{4}{c|}{	Quantitative Evaluation} & \multicolumn{3}{c}{User Study} \\
\cmidrule{2-8}          & TA $\uparrow$ & TC $\uparrow$ & LPIPS-N $\downarrow$ & LPIPS-S $\downarrow$ & Text Alignment $\downarrow$ & Content Alignment $\downarrow$ & Motion Alignment $\downarrow$ \\
\midrule
	Follow-Your-Pose \cite{ma2024follow} & 23.63  & 0.913  & 0.213  & 0.614  & \multicolumn{1}{c}{87.14\%} & \multicolumn{1}{c}{96.66\%} & \multicolumn{1}{c}{90.95\%} \\
    MotionDirector \cite{zhao2023motiondirector} & 23.90  & 0.872  & 0.141  & 0.430  & \multicolumn{1}{c}{94.55\%} & \multicolumn{1}{c}{96.66\%} & \multicolumn{1}{c}{86.18\%} \\
    Tune-A-Video \cite{wu2023tune} & 25.75  & 0.924  & 0.107  & 0.429  & \multicolumn{1}{c}{78.81\%} & \multicolumn{1}{c}{82.15\%} & \multicolumn{1}{c}{84.05\%} \\
    MotionEditor \cite{tu2023motioneditor} & 24.59  & 0.898  & 0.132  & 0.440  & \multicolumn{1}{c}{76.43\%} & \multicolumn{1}{c}{82.38\%} & \multicolumn{1}{c}{80.95\%} \\
    \midrule
    Ours  & \textbf{26.74} & \textbf{0.941}  & \textbf{0.092} & \textbf{0.363} & \multicolumn{1}{c}{-} & \multicolumn{1}{c}{-} & \multicolumn{1}{c}{-} \\
    \bottomrule
    \end{tabular}%
  
  \label{tab:comparison_wild}%
\end{table*}%

\subsection{Ablation Study}

\begin{figure}
	\centering
	\includegraphics[width=0.46\textwidth]{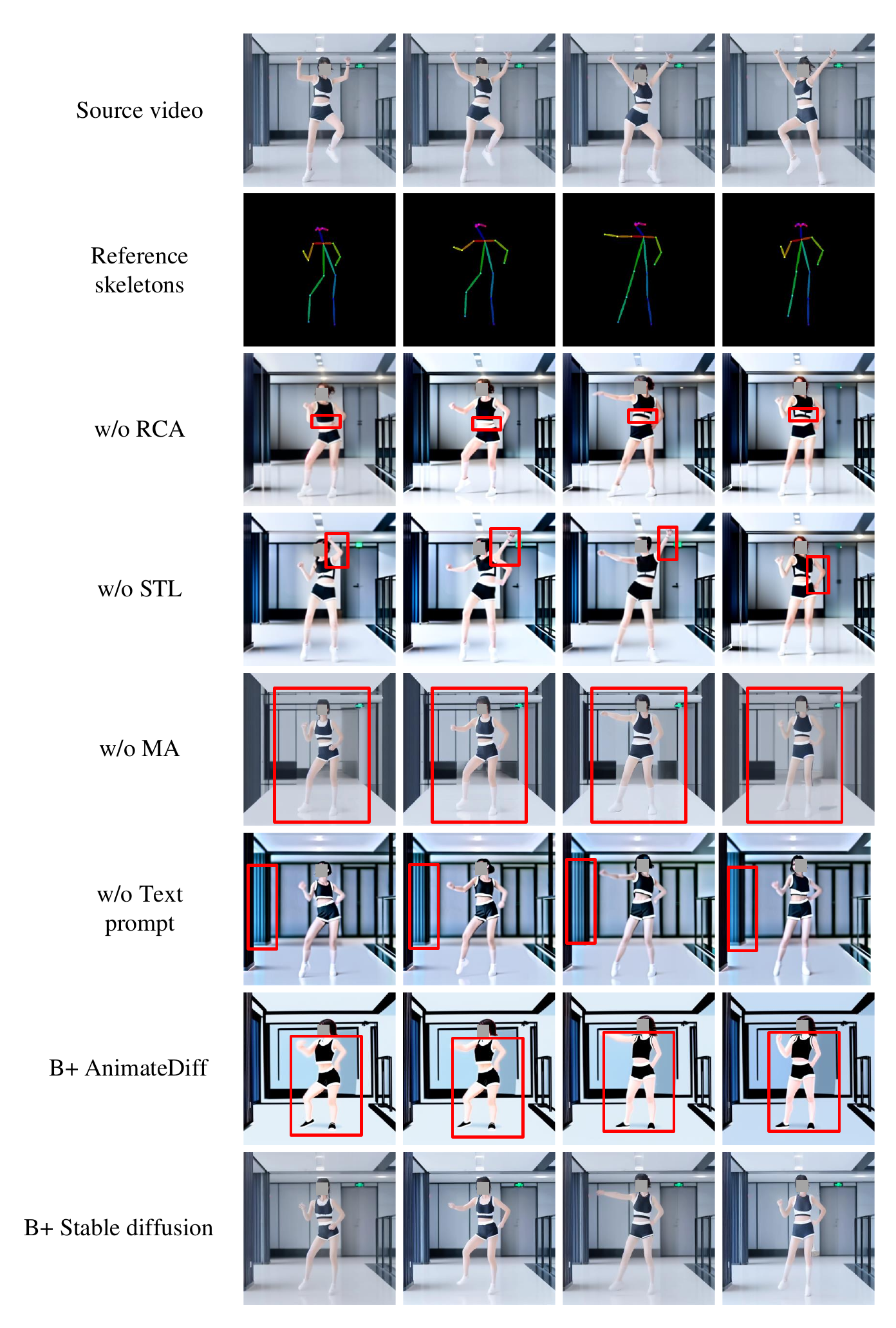}
	\caption{The Ablation Study of Edit-Your-Motion. We conducted ablation experiments on RCA, STL, MA, and text prompt individually, thereby demonstrating the importance of each module. Additionally, we used AnimateDiff and Stable Diffusion v1.5 to evaluate the effect of the backbone on the results.}
	\label{fig:Ablation2}
\end{figure}

To evaluate the effectiveness of the proposed modules, we conducted an ablation study on each component. Table \ref{tab:ablation_study} presents the quantitative results for the three modules: RCA, MA, and STL. It is evident that STL significantly improves PSNR, while MA demonstrates improved performance in SSIM and LPIPS. This improvement is attributed to STL's ability to help the network learn temporal and spatial features of the video in fewer iterations by decoupling learning. 
On the other hand, MA enhances the fusion of human and background features by addressing the conflict between skeleton and spatial features.

Additionally, we have visualized further ablation studies in Fig. \ref{fig:Ablation2}. The results indicate that RCA and text prompts contribute to maintaining appearance consistency across frames, while STL effectively aligns motion consistency between the generated video and the reference skeletons. MA helps resolve the conflict between the human and the surrounding environment. 

We utilized Stable Diffusion v1.5 (SD1.5) and AnimateDiff to evaluate the impact of the backbone on the results. Our findings indicate that SD1.5 is more effective at maintaining the consistency of appearance between the generated video and the source video compared to AnimateDiff.

\section{CONCLUSION}
In this paper, we explore the application of video motion editing on the unseen in-the-wild cases and propose Edit-Your-Motion, a method for learning temporal and spatial features separately in a spatio-temporal diffusion model.
Edit-Your-Motion avoids complexity by eliminating the need for a two-branch structure, significantly reducing GPU memory usage.
Specifically, we use DDIM inversion to maintain the appearance of the source video and introduce a motion attention module to align the motion with the reference skeleton video.
Then, we design a spatio-temporal decoupled two-stage training strategy to decouple learning spatial and temporal features with a fewer iterations.
To enhance inter-frame consistency, we replace spatial attention with recurrent causal attention, which explicitly establishes connections between frames.
In the inference stage, we invert the source images/video as latent noise. Then, skeleton and appearance features are injected into U-Net through motion attention to realize video motion editing.
Experiments on YouTube and TikTok benchmark demonstrate the robustness of our method in the unseen in-the-wild cases.
In the future, we will apply Edit-Your-Motion to animations, advertisements, short videos and other multimedia fields, so that users can create more interesting and innovative videos.

\bibliography{references}
\bibliographystyle{IEEEtran}
\end{document}